\documentclass[conference]{IEEEtran}
\IEEEoverridecommandlockouts
\usepackage{cite}
\usepackage{amsmath,amssymb,amsfonts}
\usepackage{algorithmic}
\usepackage{graphicx}
\usepackage{textcomp}
\usepackage{xcolor}
\def\BibTeX{{\rm B\kern-.05em{\sc i\kern-.025em b}\kern-.08em
    T\kern-.1667em\lower.7ex\hbox{E}\kern-.125emX}}

\begin{document}

\title{Infrequent Child-Directed Speech Is Bursty and May Draw Infant Vocalizations\\
%{\footnotesize \textsuperscript{*}Note: Sub-titles are not captured for https://ieeexplore.ieee.org  and
}

\author{\IEEEauthorblockN{Margaret Cychosz}
\IEEEauthorblockA{\textit{Dept. of Linguistics} \\
\textit{Stanford University}\\
Stanford, USA \\
mcychosz@stanford.edu}
\and
\IEEEauthorblockN{Adriana Weisleder}
\IEEEauthorblockA{\textit{Dept. of Communication Sciences and Disorders} \\
\textit{Northwestern University}\\
Evanston, USA \\
adriana.weisleder@northwestern.edu}
}

\maketitle

\begin{abstract}
Children in many parts of the world hear relatively little speech directed to them, yet still reach major language development milestones. What differs about the speech input that infants learn from when directed input is rare? Using longform, infant-centered audio recordings taken in rural Bolivia and the urban U.S., we examined temporal patterns of infants' speech input and their pre-linguistic vocal behavior. We find that child-directed speech in Bolivia, though less frequent, was just as temporally clustered as speech input in the U.S, arriving in concentrated bursts rather than spread across the day. In both communities, infants were most likely to produce speech-like vocalizations during periods of speech directed to them, with the probability of infants' speech-like vocalizations during target child-directed speech nearly double that during silence. In Bolivia, infants' speech-like vocalizations were also more likely to occur during bouts of directed speech from older children than from adults. Together, these findings suggest that the developmental impact of child-directed speech may depend not only on quantity, but on temporal concentration and source, with older children serving as an important source of input in some communities, including where adult speech to infants is less frequent.

\end{abstract}


\section{Introduction}

Decades of research conducted primarily in industrialized societies have established links between the quantity and quality of child-directed speech (CDS) and child language development: children who hear more speech directed to them develop larger vocabularies, process words more efficiently, and use more complex grammatical constructions \cite{huttenlocher_language_2002,rowe_child-directed_2008,weisleder_talking_2013}. Yet children in many communities around the world hear relatively little directed speech, often a fraction of what is typical in Western, middle-class households \cite{cristia_child-directed_2019,scaff_characterization_2024}. Nevertheless, research suggests that these children still reach major language milestones on a comparable timeline \cite{casillas_early_2020}. If CDS matters for language learning, how do children learn when it is infrequent?

Researchers have proposed a number of answers to this question: some argue it is not the quantity of directed input that matters, but the quality of exchanges \cite{hirsh-pasek_contribution_2015, rowe_longitudinal_2012}. Others show that when broader definitions of directed input are employed (encompassing not just adults but other children, overheard speech, etc.) differences in quantities of directed input between communities disappear \cite{cychosz_child-directed_nodate,scaff_characterization_2024,sperry_reexamining_2019}. Here we suggest another possibility: it is not just the \textit{amount} or the linguistic characteristics of directed speech that matters, but \textit{when} it arrives \cite{Abney2025}. 

Speech input from caregivers is not evenly distributed across children's days. It arrives in clusters and bursts separated by periods of relative quiet, the child's self-play, or electronic media consumption. Recent work has shown that bursty, temporally clustered speech input predicts vocabulary size in a broad sample of U.S. children (N=292, aged 2-7 years), even after controlling for total input quantity \cite{cychosz_bursty_2025}. This finding aligns with exposure-consolidation models of learning, which propose that concentrated doses of input followed by down periods may support encoding and retention \cite{bernier_sleep_2013, henderson_learning_2013}. If temporal clustering matters for learning, then the distribution of speech input over time may be one mechanism by which children acquire language, including in communities where directed input is relatively infrequent. 

Another, not mutually-exclusive possibility, is that the source of CDS matters. In Western communities, it is typically assumed that the primary source of speech input to children is the main caregiver (typically the mother) speaking directly to the child \cite{hoff_specificity_2003,snow_mothers_1972}. But in many communities in the world where alloparental care is common, infants are commonly surrounded by many other older children, including siblings, but also cousins and neighbors, and regularly hear speech directed to others, both children and adults \cite{ivey_cooperative_2000,weisner_socialization_1987,tronick_multiple_1987}. Studies suggest that in these settings, infants could acquire some components of language, such as a vocabulary, from overheard in addition to directed speech \cite{foushee_infants_2024}. Moreover, the CDS that infants hear could stem from children rather than adults. This last component matters because children speak differently than adults: their speech is slower, more variable, and less grammatically complex well into the school-aged years and even adolescence \cite{rice_mean_2010,lee_acoustics_1999}. If infants are sensitive to being addressed, they may engage vocally with input from other children just as they do with input from adults, or perhaps more so if that input from children were more accessible.

In this study, we examine the temporal distribution and sources of CDS in longform, infant-centered recordings taken in two communities: a rural Indigenous Quechua- and Spanish-speaking community in southern Bolivia and an urban Spanish- and English-speaking immigrant community in the U.S. Previous work established that infants in this community in Bolivia hear far less speech directed specifically to them than those in the U.S. community, but comparable amounts of CDS overall once speech to other children nearby is included \cite{cychosz_child-directed_nodate}. We follow up on that work and pose three questions: (1) Is CDS more temporally clustered in Bolivia, where it is rarer? (2) Do infants in both communities produce more speech-like vocalizations when they hear speech directed right to them than when they hear speech directed to other children or adults, even if that directed speech is less frequent overall? (3) Does the source of CDS to the target infant, stemming from an adult versus an older child, predict the occurrence of infants' speech-like vocalizations across these two communities?

\section{Methods}

\subsection{Participants}

Participants were infants from a rural bilingual Quechua- and Spanish-speaking community in the southern highlands of Bolivia (N=5, 3 female and 2 male; 5.7–12.3 months, M=8.66, SD=2.79) and from bilingual Spanish- and English-speaking immigrant families in the urban U.S. (N=5, 2 female and 3 male; 6.5–12.6 months, M=8.94, SD=2.54). Infants were matched as closely as possible across communities by age, gender, and number of siblings. Bolivian infants still had more siblings (mode=2) than U.S. infants (mode=1), reflecting typical family structures in these communities. All U.S. caregivers identified as Hispanic or Latino and were first-generation immigrants to the U.S. from Latin America.

\subsection{Data Collection}

Each infant wore a small, lightweight (2''x3''; 2 oz.) Language ENvironment Analysis audio recorder in a special pocket sewed to the front of a vest or shirt for an entire day, capturing the infant's full auditory environment within approximately a 10-foot radius. Families were instructed to go about their normal daily activities. Recordings from Bolivia averaged 10.47 hours (SD = 3.23; range: 7.77–16.0); all U.S. recordings were 16 hours.

\subsection{Data Annotation}

Recordings were annotated by the authors and trained research assistants. First, sleep periods and periods when a researcher was present were manually excluded from analysis. Then, every other 30-second clip was manually annotated (+/- present) for 1) \textsc{addressee}: speech directed \textit{just} to the target infant, speech directed to \textit{any} child including the target infant, and speech directed to adults, and 2) \textsc{speaker}: adult female, adult male, other child (in all cases the other child was older than the infant). 

This analysis permitted a 3x4 coding scheme crossing speaker (female adult, male adult, older child) with addressee (target infant, other child, adult, unsure) and allowed us to code the following main speech registers (1) \textit{target child-directed speech} (TCDS): speech directed just to the target infant, by an adult or older child (2) \textit{inclusive child-directed speech} (CDS incl.): speech directed to any child within the target infant's vicinity, including the target infant, by an adult or older child, (3) \textit{exclusive child-directed speech} (CDS excl.): speech to any child other than the target infant, and (4) \textit{adult-directed speech} (ADS): speech spoken by an adult to another adult. When speech was present but the intended addressee could not be determined from the surrounding context (e.g. lexical content), the addressee was coded as \textit{unsure}. From these categories we additionally derived \textit{overheard speech}, a category that includes speech to other children and adults but not the target infant, and report descriptive statistics on this category in results. Within each 30-second clip, annotators also binarily coded  (+/-present) whether the target infant produced a speech-like vocalization characterized as a fully-resonant vowel (e.g. ``eee'', ``ooo'') or vocalization with a consonant-vowel transition (e.g. ``ba'', ``am'') \cite{oller_emergence_2000}, but excluding cry and vegetative sounds such as feeding. Inter-rater reliability was high (Cohen's $\kappa$ $>$ .82 for all speaker/addressee combinations; see \cite{cychosz_child-directed_nodate} for details).

\subsection{Computation of burstiness}

Temporal clustering of speech was quantified using the burstiness parameter $B$ \cite{goh_burstiness_2008}, computed from inter-event intervals: $B = \frac{\sigma - \mu}{\sigma + \mu}$ where $\mu$ is the mean inter-event interval and $\sigma$ is the standard deviation of those intervals, here measured as the number of 30-second segments between consecutive segments containing a given speech register. $B$ ranges from -1 (indicating that speech input is perfectly regularly spaced throughout the day) to +1 (indicating that speech input is maximally clustered), with $B$ = 0 indicating random timing of speech input. 

\section{Results}

\subsection{Is child-directed speech more temporally clustered (burstier) in Bolivia, where it is rarer?} 

\begin{figure*}
\centering
\includegraphics[width=\textwidth]{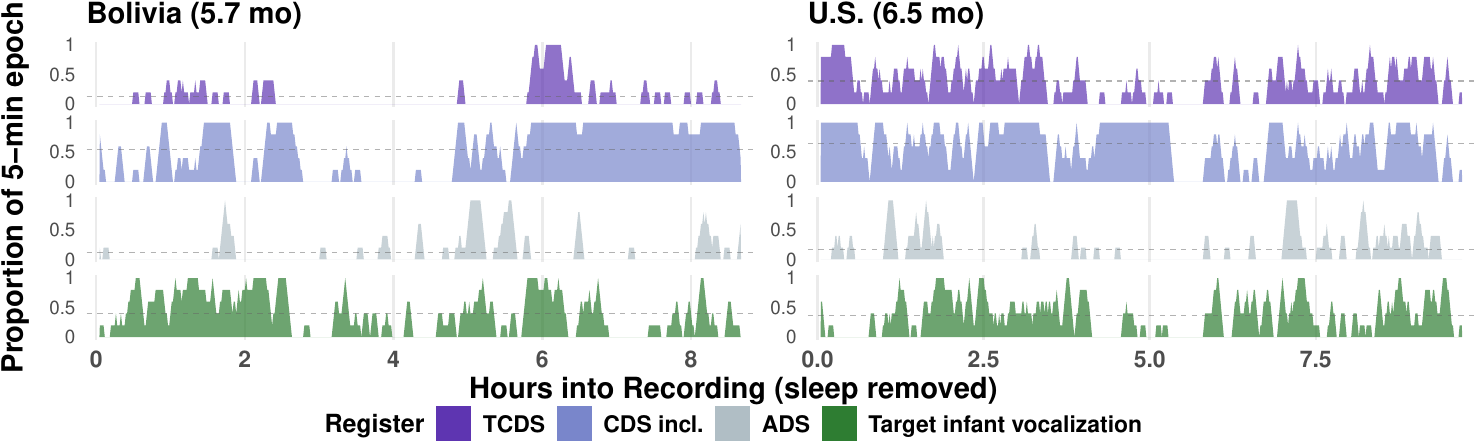}
\caption{Daily speech register patterns for two representative infants in Bolivia (left) and the U.S. (right). Y-axis=the proportion of a 5-minute rolling window containing a given register. Dashed line indicates the mean proportion of each register or vocalization rate over the entire recording period.}
\label{fig:time-series}
\end{figure*}

Raw burstiness metrics suggest community-level differences, with TCDS nearly twice as bursty in Bolivia (M=0.27, SD=0.16) as the U.S. (M=0.15, SD=0.11) while ADS was more comparable (Table \ref{tab:register-stats}). However, TCDS is also rarer in the Bolivian infants' environments (occurring in just approximately 12\% of segments compared to 39\% for the U.S.; Figure \ref{fig:time-series}). Given these baseline differences, we computed a permutation null distribution for each infant by register combination by shuffling the binary time series (+/- register present) for each infant by register combination 1,000 times and recalculating $B$. This baseline correction allowed us to determine whether the burstiness exceeded what would be expected by chance given the overall prevalence of each register for each infant, and was critical for the true cross-community comparison given lower overall amounts of TCDS in Bolivia. 

To then evaluate community-level differences, we fit a linear mixed effects model with baseline-corrected $B$ (burstiness) as the dependent variable and community (Bolivia, U.S.), register (TCDS, CDS\_excl, ADS, a non-overlapping set to avoid multicollinearity in the model), and their interaction as fixed effects, with a random intercept for each infant. Results show no reliable effect of community, register, or their interaction (all \textit{p}s $>$ .05). Finally, because our key question focused on examining community-level differences in the temporal structure of TCDS in particular, we fit another linear mixed effects model with baseline-corrected $B$ of TCDS as the dependent variable, and community (Bolivia, U.S.) as a fixed effect. Baseline-corrected TCDS was not significantly more bursty in Bolivia ($M$ = 0.31, $SD$ = 0.15) than the U.S. ($M$ = 0.27, $SD$ = 0.09; $\beta$ = $-$0.04, $t$(8) = $-$0.49, $p$ = .635), suggesting that the temporal distribution (i.e. predictability) of TCDS is statistically equivalent between communities once these baseline frequency rates are taken into account.  

\begin{table}
\caption{Burstiness ($B$) prior to infant-level baseline correction, by register and community.\label{tab:register-stats}}
\begin{center}
\begin{tabular}{|l|l|l|l|}
\hline
\textbf{Community} & \textbf{Register} & \textbf{M (SD)} & \textbf{Range} \\
\hline
Bolivia & TCDS\textsuperscript & 0.27 (0.16) & 0.09--0.46 \\
        & CDS incl.\textsuperscript & 0.22 (0.10) & 0.13--0.38 \\
        & CDS excl.\textsuperscript & 0.34 (0.13) & 0.14--0.47 \\
        & ADS\textsuperscript & 0.37 (0.17) & 0.13--0.59 \\
        & Overheard\textsuperscript & 0.21 (0.12) & 0.07--0.31 \\
\hline
U.S.    & TCDS\textsuperscript & 0.15 (0.11) & 0.04--0.31 \\
        & CDS incl.\textsuperscript & 0.10 (0.11) & $-$0.06--0.23 \\
        & CDS excl.\textsuperscript & 0.20 (0.26) & $-$0.06--0.57 \\
        & ADS\textsuperscript & 0.40 (0.05) & 0.35--0.45 \\
        & Overheard\textsuperscript & 0.27 (0.21) & $-$0.06--0.47 \\
%\hline
%\multicolumn{4}{|l|}{\textsuperscript{a}Target child-directed speech.} \\
%\multicolumn{4}{|l|}{\textsuperscript{b}Directed to any child including the target infant.} \\
%\multicolumn{4}{|l|}{\textsuperscript{c}Directed to any child except the target infant.} \\
%\multicolumn{4}{|l|}{\textsuperscript{d}Adult-directed speech.} \\
%\multicolumn{4}{|l|}{\textsuperscript{e}Speech directed to any adult or child except the target infant.} \\
\hline
\multicolumn{4}{l}{%
  \parbox{\dimexpr\columnwidth-2\tabcolsep\relax}{%
    \vspace{2pt}
    \footnotesize TCDS = speech directed to the target infant only. CDS incl.\ = child-directed speech to the target child and other children. CDS excl.\ = child-directed speech to other children only. ADS = adult-directed speech.
    \vspace{2pt}
  }%
}
\end{tabular}
\end{center}
\end{table}

\subsection{Do infants in both communities produce more speech-like vocalizations when they hear speech directed right to them than when they hear speech directed to other children or adults, even if that directed speech is less frequent overall?} 

\begin{figure*}
\centering
\includegraphics[width=\textwidth]{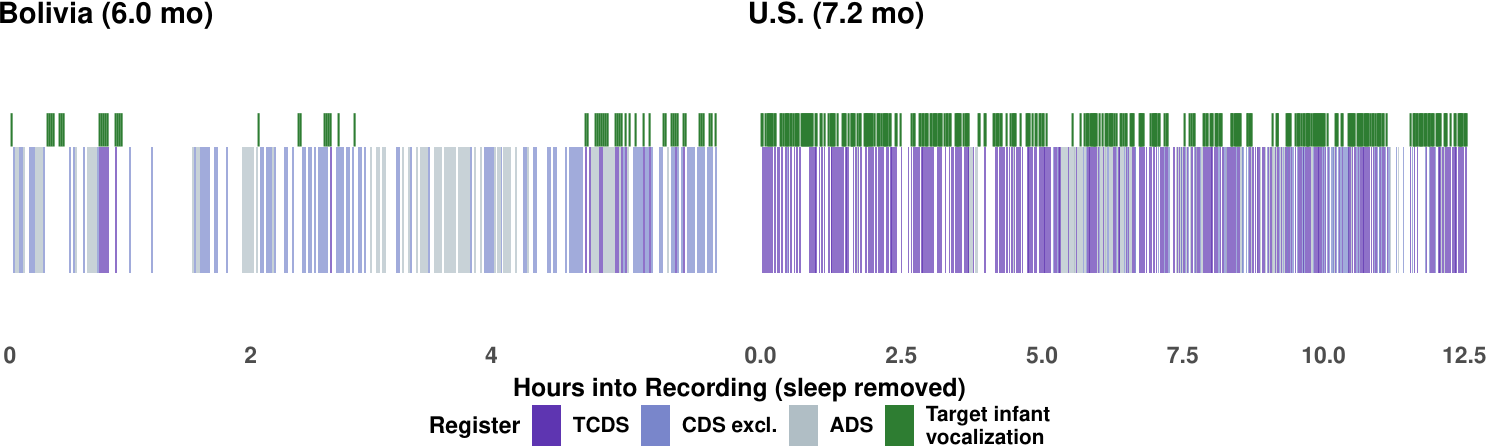}
\caption{Temporal distribution of speech registers and target infant vocalizations for two representative infants in Bolivia (left) and the U.S. (right). Each vertical slice represents the presence or absence of the register/vocalization within a single 30-second annotation bin. White space indicates silence (no speech present in the segment). Green tick marks above the register bar indicate bins containing a speech-like vocalization from the target infant.}
\label{fig:raster}
\end{figure*}

\begin{figure}
\centering
\caption{Conditional probability of target infant vocalization given each speech register, by community. Points represent individual infants. Vocalizations are more likely to co-occur with TCDS than most other registers, in both communities (see Table 2 for detail). TCDS = speech directed to the target infant and no other child. CDS incl. = child-directed speech to the target child and other children. CDS excl. = child-directed speech to other children only. ADS = adult-directed speech.}
\label{fig:boxplot}
\includegraphics[width=.48\textwidth]{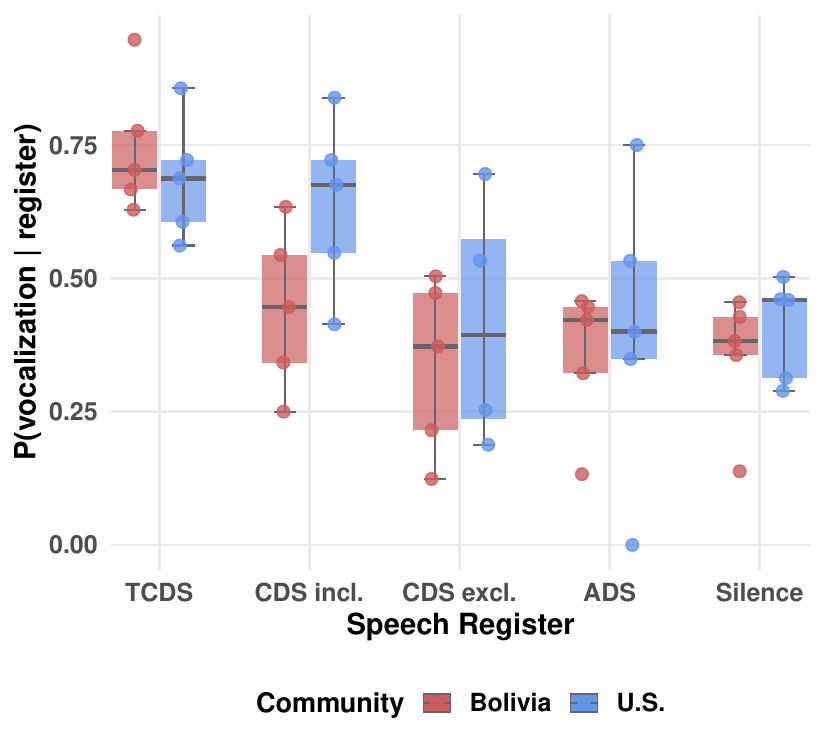}
\end{figure}

For this analysis, we compute the conditional probability of an infant vocalizing within the same 30-second segment as a given speech register (Figure \ref{fig:raster}). Descriptive statistics show lower probabilities of infant vocalization as the speech becomes less ``infant-centered'' (i.e. TCDS $>$ CDS incl. $>$ CDS excl. $>$ ADS $>$ silence; Table \ref{tab:voc-probability}). To evaluate if the probability of infant vocalization reliably differed by register, we conducted a series of paired, one-tailed t-tests within each community (four pairwise, corrected comparisons). Results show significantly higher probabilities of infants vocalizing in response to TCDS than most other speech registers that do not contain TCDS (corrected \textit{p}s $<$ .05; Table \ref{tab:voc-probability}) in both the U.S. and Bolivian communities (Figure \ref{fig:boxplot}). 

\begin{table}
\centering
\caption{P(vocalization $|$ register) and one-tailed paired $t$-tests comparing TCDS to each register, by community. $p$-values are Bonferroni-corrected (4 comparisons per community). Bold indicates significantly higher probability of vocalization in TCDS than the listed register, after correction. \label{tab:voc-probability}}
\centering
\fontsize{9}{11}\selectfont
\begin{tabular}[t]{|l|c|c|}
\hline
  & Bolivia & U.S.\\
\hline
\multicolumn{3}{|l|}{\textbf{P(voc $|$ register)}}\\
\hline
\hspace{1em} & M (SD) & M (SD)\\
\hspace{1em}Target child-directed speech & 0.74 (0.13) & 0.69 (0.11)\\
\hspace{1em}Child-directed speech incl. & 0.44 (0.15) & 0.64 (0.16)\\
\hspace{1em}Child-directed speech excl. & 0.34 (0.16) & 0.42 (0.24)\\
\hspace{1em}Adult-directed speech & 0.36 (0.14) & 0.41 (0.28)\\
\hspace{1em}Silence & 0.35 (0.13) & 0.4 (0.1)\\
\hline
\multicolumn{3}{|l|}{\textbf{TCDS vs. other registers}}\\
\hline
\hspace{1em} & $t$(4), $p$ & $t$(4), $p$\\
\hspace{1em}TCDS vs. CDS incl. & 2.90, 0.088  & 1.75, 0.312 \\
\hspace{1em}TCDS vs. CDS excl. & 3.64, \textbf{0.044} & 4.37, \textbf{0.044}\\
\hspace{1em}TCDS vs. ADS & 3.61, \textbf{0.045} & 2.68, 0.111\\
\hspace{1em}TCDS vs. Silence & 3.65, \textbf{0.044} & 8.9, \textbf{0.002}\\
\hline
\multicolumn{3}{l}{%
  \parbox{\dimexpr\columnwidth-2\tabcolsep\relax}{%
    \vspace{2pt}
    \footnotesize TCDS = speech directed to the target infant only. CDS incl.\ = child-directed speech to the target child and other children. CDS excl.\ = child-directed speech to other children only. ADS = adult-directed speech. One U.S. infant did not have sufficient ADS in the recording to contribute to the analysis of ADS.
    \vspace{2pt}
  }%
}
\end{tabular}
\end{table}

\subsection{Does the source of child-directed speech to the target infant, stemming from an adult versus an older child, predict the occurrence of infants' speech-like vocalizations across these two communities?}

We again assessed the probability of infant vocalization co-occurrence within 30-second segments, but now measuring it with TCDS spoken by older children versus adults (male and female). We conducted paired one-tailed t-tests comparing the probability of infant vocalization in each context. Results showed that Bolivian infants were significantly more likely to vocalize in response to speech directed to them from older children than from adults ($t$(4) = 2.92, $p$ = .043), while there were no differences in the probability of infant vocalization in response to speech directed to them from older children versus adults for the U.S. infants ($t$(3) = 1.54, $p$ = .222).

\section{Discussion}

This paper examined the temporal distribution of speech in longform, infant-centered recordings taken in two communities: a rural Indigenous Quechua- and Spanish-speaking community in southern Bolivia and an urban Spanish- and English-speaking immigrant community in the U.S. Previous work showed that infants in this Bolivian community hear less speech directed to them than those in the U.S. Here, quantifying burstiness \cite{goh_burstiness_2008}, we find that TCDS is more temporally clustered in Bolivia than in the U.S. However, when correcting for baseline differences in the frequency of TCDS, we no longer find differences between communities. This result suggests that TCDS may be more temporally clustered in the Bolivian community \textit{because} it is less frequent, allowing for longer periods of time between bursts. In contrast, in the U.S. community, TCDS occurs more frequently throughout the day, resulting in a more uniform distribution. We also examined how the type and source of speech available to the infants impacted the likelihood of infants' co-occurring vocalizations. We found that infants were more likely to produce speech-like vocalizations during periods where they heard speech directed to them than during periods of silence or those containing speech directed to other children or adults. In addition, in Bolivia but not in the U.S., infants were more likely to vocalize during periods with TCDS from older children than during periods with TCDS from adults. In the following sections we discuss each of these findings and how they relate to models of language acquisition. 

\subsection{Child-directed speech is burstier in the Bolivian community, where it is rarer}

Research on the role of input in language acquisition has traditionally focused on the amount and characteristics of speech available to infants \cite{rowe_analyzing_2020, rowe_longitudinal_2012, hart_meaningful_1995}. However, there is increasing recognition that the temporal structure of language input, or the distribution of speech over time, could also impact infants' ability to learn \cite{mendoza_everyday_2021, Abney2025, de_barbaro_ten_2022}. Exposure-consolidation models of learning propose that concentrated bursts of input followed by down periods may support encoding and retention of linguistic information \cite{bernier_sleep_2013, henderson_learning_2013}. This hypothesis has been proposed as one explanation for the seemingly puzzling finding that children in communities in which directed speech is rare nevertheless develop language on a comparable timeline as those with greater access to directed input \cite{casillas_early_2020, casillas_early_2021, foushee_infants_2024}. If children benefit from experiencing bursts of linguistic input followed by lulls, then the temporal clustering of CDS might be particularly important in communities with little directed speech.  

This hypothesis motivated our analysis of the temporal clustering of CDS in two communities differing in TCDS frequency \cite{cychosz_child-directed_nodate}. Across both communities, burstiness estimates ranged from .04 to .46 across individual infants. Since burstiness values $>$0 indicate that speech is bursty (as opposed to regularly spaced), with 1 indicating that speech is maximally clustered, our estimates indicate that TCDS was moderately bursty in both communities. Because this is the first study applying the burstiness metric to the distribution of CDS in longform recordings, we are not able to compare the magnitude of our burstiness estimates to other studies, but these values can now provide a useful benchmark for studies with other age groups and populations. 

We found that TCDS was burstier in the Bolivian community than in the U.S. Yet these differences disappeared when equating for the frequency of TCDS across communities. This result raises interesting questions about the relationship between the frequency and temporal structure of language input. One recent study examined the relationship between burstiness and frequency of individual words and found that, while related, word burstiness is not a redundant measure of word frequency, as words that are bursty can have a wide range of frequencies \cite{Marasli2025}. We propose something similar with respect to the relationship between frequency and burstiness of CDS. Although the two are distinct constructs that can have unique effects on language learning, indeed a recent study found that bursty speech input predicted vocabulary among U.S. children after controlling for input quantity \cite{cychosz_bursty_2025}, they are not completely independent. In future work, it will be important to understand if and how frequency and burstiness condition each other across different contexts. Understanding differences in burstiness would also benefit from a deeper examination of the processes that generate bursts of language interaction across an infant's day (e.g., activities, sleep-wake rhythms).  

\subsection{Infants vocalize more in the presence of directed speech, including from older children}

A prominent theory of early speech development posits that vocal feedback loops between infants and their caregivers are key contributors to vocal development, including early phonological outcomes like babbling maturity \cite{warlaumont_social_2014,goldstein_social_2008,bornstein_mother-infant_2015}. Previous studies have found that infants are more likely to produce speech-related vocalizations when they receive an adult response to a previous vocalization, as well as in response to maternal vocalizations. However, a majority of these studies focused on caregiver-infant dyads in societies where one-on-one verbal exchanges between parents and infants are common. What has been less clear is whether infants are more likely to vocalize in response to different types of speech, such as directed versus overheard, and from different sources. 

Here, we find that infants in both the Bolivian and U.S. communities were more likely to produce speech-like vocalizations in the presence of speech that was directed to them than during periods of silence. Interestingly, infants were also more likely to vocalize in the presence of speech that was directed to them than in the presence of speech directed to others, with no reliable differences in the likelihood of vocalizations between speech directed to other children versus adults. This result suggests that even in contexts in which directed speech is infrequent, as in the Bolivian community, this type of speech may still more effectively elicits infant vocalizations. An important caveat of this conclusion, however, is that the vocalization co-occurrence we measure captures whether an infant vocalization and a speech register occur within the same 30-second segment, not necessarily whether the vocalization is temporally contingent with speech. So from these data we cannot conclude that an infant's vocalization was necessarily produced in response to the speech in that segment, or whether it occurred independently \cite{long_social_2020}. Ongoing work we are conducting using finer-grained temporal contingencies will allow us to examine these patterns in more detail, and make more definitive conclusions about infant vocalization patterns in response to ambient speech. Nevertheless, although we did not examine the precise timing of vocalizations and cannot infer the presence of vocal feedback loop, these findings are in line with those from previous studies suggesting that contingent vocal feedback from caregivers can be an important driver of infant vocalizations. 

Finally, we found that Bolivian infants, but not U.S. infants, were more likely to vocalize when the speech directed to them was from older children rather than from adults. Infants in this Bolivian community tend to have a larger number of siblings than those in the U.S. community, and young children are nearly as likely to socialize with neighboring children as their own siblings. Thus, this finding suggests that differences in children's socialization environments, particularly the much higher presence of additional, older children in the Bolivian community, influence the sources of language input that infants are most likely to learn from. Importantly, this finding also reinforces the need to broaden our methodological and analytical frames when studying the role of input in language development, including not only child-directed and overheard speech, as has become more common, but also speech from other children \cite{bulgarelli_quantifying_2022}. 

\section*{Acknowledgment}

The authors thank the families who participated in this research, as well as Anele Villanueva and Alan Mendelsohn for support in data collection and analysis. Additional thanks to Jan Edwards for the use of her recording equipment. 

%\section*{References}

\bibliographystyle{IEEEtran}
\bibliography{references-3}

\end{document}